\def\paperlanguage{}
\pdfoutput=1 

\documentclass[letterpaper, 10 pt, conference]{ieeeconf}  

\usepackage{bm}
\usepackage{cite}
\usepackage{flushend}
\include{preamble}

\newcommand{\ctext}[1]{\raise0.2ex\hbox{\textcircled{\scriptsize{#1}}}}

\IEEEoverridecommandlockouts                              

\title{\LARGE \textbf
  {
    \switchlanguage%
    {%
      Binary State Recognition by Robots using Visual Question Answering of Pre-Trained Vision-Language Model
    }%
    {%
      事前学習済みVQAモデルを用いたロボットによる状態認識
    }%
  }
}

\author{Kento Kawaharazuka$^{1}$, Yoshiki Obinata$^{1}$, Naoaki Kanazawa$^{1}$, Kei Okada$^{1}$, and Masayuki Inaba$^{1}$
  \thanks{$^{1}$ The authors are with the Department of Mechano-Informatics, Graduate School of Information Science and Technology, The University of Tokyo, 7-3-1 Hongo, Bunkyo-ku, Tokyo, 113-8656, Japan.
    {\texttt\small [kawaharazuka, obinata, kanazawa, k-okada, inaba]@jsk.t.u-tokyo.ac.jp}
  }
}
\begin{document}

\maketitle
\thispagestyle{empty}
\pagestyle{empty}

\begin{abstract}
  \switchlanguage%
  {%
    Recognition of the current state is indispensable for the operation of a robot.
    There are various states to be recognized, such as whether an elevator door is open or closed, whether an object has been grasped correctly, and whether the TV is turned on or off.
    Until now, these states have been recognized by programmatically describing the state of a point cloud or raw image, by annotating and learning images, by using special sensors, etc.
    In contrast to these methods, we apply Visual Question Answering (VQA) from a Pre-Trained Vision-Language Model (PTVLM) trained on a large-scale dataset, to such binary state recognition.
    This idea allows us to intuitively describe state recognition in language without any re-training, thereby improving the recognition ability of robots in a simple and general way.
    We summarize various techniques in questioning methods and image processing, and clarify their properties through experiments.
  }%
  {%
    ロボットの動作には現在の状態認識が欠かせない.
    エレベータや自動ドアの開閉状態, 物を正しく受け渡すことができたかどうか, 天気や電気の状態など, 認識したい状態は様々である.
    これまでこれらの状態認識は, 人間がポイントクラウドの状態をプログラムで記述, 画像についてアノテーションを行い学習, または別のセンサにより状態を検知といった方法で行われてきた.
    これらに対して本研究では, 大規模なデータセットを用いて訓練された事前学習済み視覚-言語モデルにおけるVisual Question Answering (VQA)を応用して状態認識を行う.
    今日の大規模モデルの発展を利用し, 一切の再学習等は行わず, ロボットにおける状態認識記述を直感的に言語で可能とすることで, 簡易かつ汎用的にロボットの認識能力が向上する.
    その際の質問方法や画像の加工に関する様々な工夫をまとめ, 実験からその性質を明らかにする.
  }%
\end{abstract}

\section{INTRODUCTION}\label{sec:introduction}
\switchlanguage%
{%
  When a robot performs a task, recognition of its surroundings, including the environment, target objects, and itself, is indispensable.
  For example, when considering the task of patrolling a house, it is necessary to check various states such as whether the refrigerator door is open, whether unnecessary electricity is turned on, whether water is left running, whether the kitchen is tidy, and whether the garbage is full \cite{okada2006tool}.
  In addition, when considering the task of delivery, the robot also needs to recognize whether the delivery is grasped correctly, whether the door is open or closed, and so on \cite{saito2011subwaydemo}.

  So far, these state recognitions have been done by directly processing images and point clouds programmatically \cite{chin1986recognition, quintana2018door}, annotating datasets for training \cite{li2020modifiedyolov3}, or by attaching a new sensor \cite{takahata2020coaxial}.
  While it is of course possible to create a program with a recognition rate of 100\% with several adjustments, these methods can only be applied to objects or environments for which data has been obtained or adjusted by a human programming, and the adaptability is low.
  In addition, the number of required recognizers increases as the number of states to be recognized increases, so resource management also becomes a problem.

  Therefore, in this study, we perform state recognition using a vision-language model trained on a large-scale dataset \cite{li2022largemodels, radford2021clip}.
  By directly using pre-trained models obtained from a wide variety of data from around the world, the model can be quickly and easily applied to a wide variety of environments, objects, etc. without re-training using datasets from the current environment.
  Among the tasks that can be performed with the Pre-Trained Vision-Language Model (PTVLM), we focus on Visual Question Answering task (VQA) \cite{antol2015vqa} that asks a question to an image and obtains the answer.
  VQA is basically a technology completed in the field of computer vision, and there are few examples of its practical use as a state recognizer in actual robots.
  In a related work, \cite{das2018eqa} has proposed a problem setting called Embodied Question Answering, in which a robot searches for answers to questions in a 3D simulation space.
  Also, \cite{ahn2022saycan} successfully combines a language model and a visual-language model to perform various tasks from language instructions.
  However, in most of these cases, a new network is constructed and trained specifically for the task, and the VQA task in PTVLM is not directly applied as a state recognizer to robots.
  Therefore, we use this VQA directly for binary state recognition in robots.
  This makes it possible for a single recognizer to determine whether a door is open or closed, whether the power is on or off, and even whether water is running or not.
  In addition, this method enables more intuitive conditional branching by language, and also facilitates the adjustment of the recognizer.
  This is a very simple idea but a new concept, and the methodology has not been established.
  In this study, we will examine how the application of VQA can revolutionize the conventional state recognition of robots by conducting several experiments, as well as confirm the properties of this recognizer.
  This paper represents a more experimental study than \cite{kawaharazuka2023ofaga}.
}%
{%
  ロボットがタスクを行う際, 環境や物体, 自身を含めた周囲の状態認識は欠かせない.
  例えば家の中の見回りタスクを考えた時, キッチンの近くで火はついていないか, 冷蔵庫のドアは開いていないか, 無駄な電気はついていないか, 水は出しっぱなしではないか, キッチンは整理整頓されているか, ゴミはいっぱいかなど, 様々な状態を確認する必要がある\cite{okada2006tool}.
  また, 配達タスクを考えた時, 正しく配達物が受け渡されたか, 部屋に入れるようにドアは開いているか, エレベータのドアは開いたかなどの状態認識も必要になる\cite{saito2011subwaydemo}.

  これまで, これらの状態認識は人間が直接画像や点群をプログラムで処理して行う\cite{chin1986recognition, quintana2018door}, アノテーションを行いデータセットを作成して学習する\cite{li2020modifiedyolov3}, 新しく別のセンサを取り付け検知する\cite{takahata2020coaxial}等, 個別のケースに対して個別の手法が取られてきた.
  そのため, もちろん調整次第で認識率100\%のプログラムを作ることができる一方, この方法ではデータを取った, または人間がプログラム上で調節した物体や環境にしか同様の認識器を適用できず, その適応性は低い.
  また, 認識したい状態が増えるだけ認識器も増加し, そのリソース管理にも問題が生じる.

  そこで本研究では, 視覚と言語に関する大規模なデータセットを用いて訓練された視覚-言語モデル\cite{li2022largemodels, radford2021clip}による状態認識を行う.
  現在の環境におけるデータセットを使った再学習等はせず, 世界中の多様なデータから得られた事前学習済みモデルを直接用いることで, 多様な環境・物体等に素早く容易に適用可能である.
  この事前学習済み視覚-言語モデル(Pre-Trained Vision-Language Model, PTVLM)が可能なタスクの中でも, 本研究ではVisual Question Answering (VQA) \cite{antol2015vqa}に着目する.
  VQAは画像に対して質問をし, その回答を得るタスクである.
  一方で, VQAは基本的にComputer Visionの分野で完結した技術であり, 実際のロボットにおける認識器として実運用される例は少ない.
  関連研究として, \cite{das2018eqa}はEmbodied Question Answeringという問題設定を打ち出し, 3Dシミュレーション空間で, ロボットが質問の答えを探索する経路計画を行っている.
  また, \cite{ahn2022saycan}は言語モデルと視覚-言語モデルを組み合わせて多様な言語instructionから多様なタスクを行うことに成功している.
  しかし, これらはそのタスク専用に新しくネットワークを構築・学習させる場合がほとんどであり, PTVLMにおけるVQAタスクを直接ロボットの認識器として組み込むものではない.
  本研究は, このVQAを直接ロボットにおける二値の状態認識に用いる.
  これにより, ドアの開閉状態や電源のオンオフ, 水が流れているかどうかまでを, 単一の認識器で行うことができるようになる.
  また, 言語によるより直感的な条件分岐が可能となり, 認識器の調整も容易になる.
  本手法は, 非常にシンプルな考え方ではあるが, この考え方は新しく, 性能の確認は行われてきていない.
  本研究では, VQAを応用することで, これまでのロボットの状態認識が如何に革新されるかを, いくつかの実験から検証し, その性質を確認する.
}%

\begin{figure}[t]
  \centering
  \includegraphics[width=1.0\columnwidth]{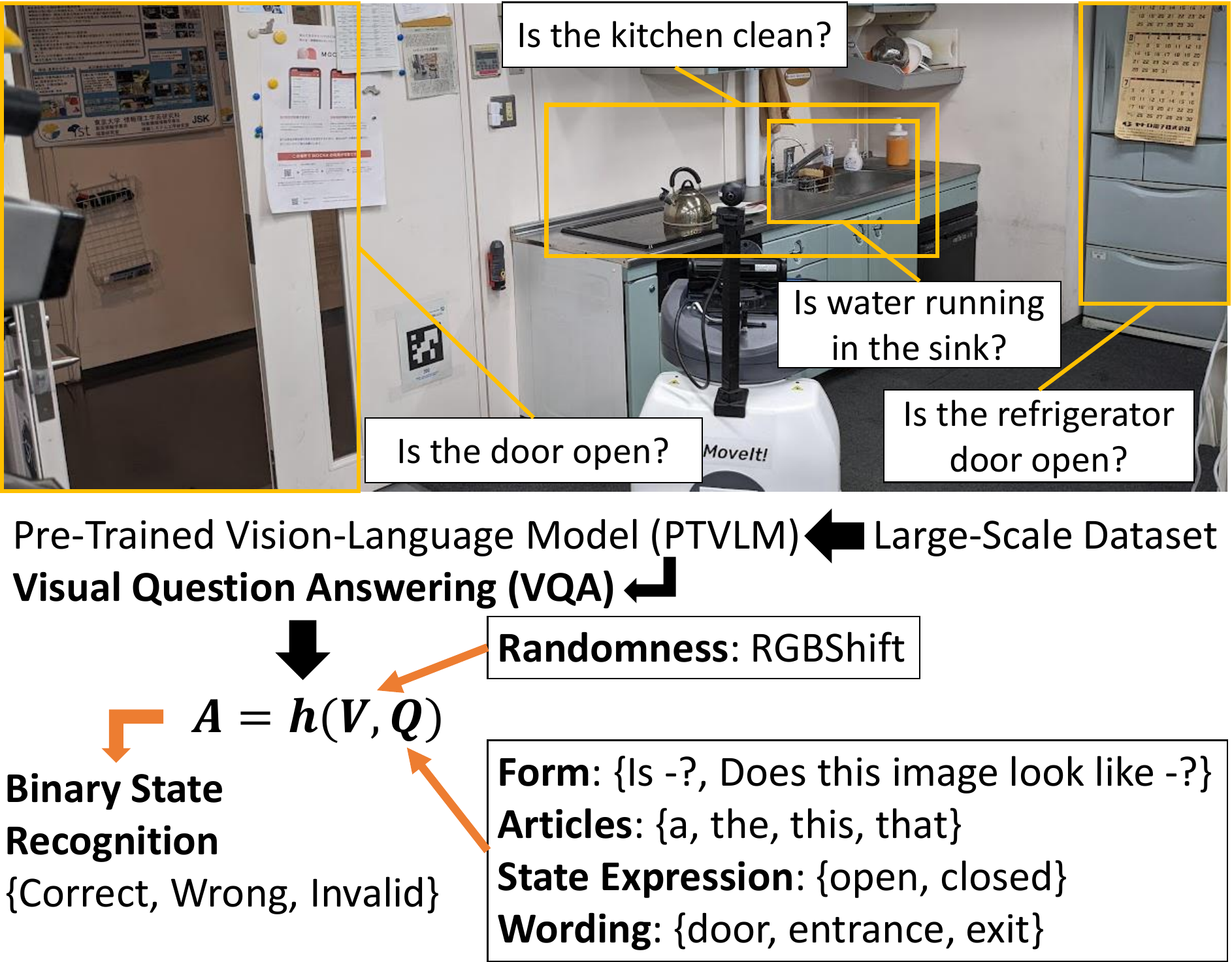}
  \vspace{-1.0ex}
  \caption{The concept of this study. The robot can execute binary state recognition using visual question answering of pre-trained vision-language model with various images and ways to express the question.}
  \label{figure:concept}
  \vspace{-3.0ex}
\end{figure}


\section{Binary State Recognition Using Pre-Trained Vision-Language Models} \label{sec:proposed}

\subsection{Pre-Trained Vision-Language Models for VQA} \label{subsec:vqa-network}
\switchlanguage%
{%
  In this study, we conduct experiments using OFA \cite{wang2022ofa} as an example of PTVLM.
  OFA is a model that can be trained to perform various tasks such as Image Captioning (IC), Visual Question Answering (VQA), and Visual Grounding (VG) in a single model.
  We mainly use VQA and a part of IC in our experiments.
  The following equation can express the relationship between the network inputs and outputs.
  \begin{align}
    \bm{A} = \bm{h}(\bm{V}, \bm{Q}) \label{eq:vqa}
  \end{align}
  where $\bm{A}$ (Answer) is a sentence, $\bm{V}$ (Vision) is an image, $\bm{Q}$ (Question) is a sentence, and $\bm{h}$ is a neural network.
  For example, for an image $\bm{V}$ of a man with glasses sitting on a chair, inputting $\bm{Q}$ ``What does the image describe?'' returns $\bm{A}$ ``A man with glasses is sitting on a chair''.
  This study makes the best use of this VQA for state recognition.
}%
{%
  本研究ではPre-Trained Vision-Language Model (PTVLM)の一例としてOFA \cite{wang2022ofa}を用いて実験を行う.
  OFAはImage Captioning (IC)やVisual Question Answering (VQA), Visual Grounding (VG)など様々なタスクを単一のモデル実行可能な形で学習されたなモデルである.
  本研究ではこの中でも主にVQAと一部ICを用いて実験を行う.
  基本的に, 以下のような式でネットワーク入力と出力の関係を表現できる.
  \begin{align}
    \bm{A} = \bm{h}(\bm{V}, \bm{Q}) \label{eq:vqa}
  \end{align}
  ここで, $\bm{A}$をAnswerとなる文章, $\bm{V}$をVisionとなる画像, $\bm{Q}$をQuestionとなる文章, $\bm{h}$をニューラルネットワークとする.
  例えば, メガネをかけた人間が椅子に座っている画像$\bm{V}$に対して, "What does the image describe?"という$\bm{Q}$を入力すると, "A man with glasses is sitting on a chair."という答えが返ってくる.
  本研究はこれを最大限に応用することで状態認識を行う.
}%

\subsection{Binary State Recognition Using VQA} \label{subsec:state-recognition}
\switchlanguage%
{%
  It is difficult to describe the conditional branching of a robot motion if the output of a state recognizer is free-form such as ``Cat'' or ``A man with glasses is sitting on a chair''.
  Therefore, in this study, we focus on state recognition specifically for robots, and handle only binary state recognition answered by Yes or No.
  If the environment has three states, we only need to execute binary state recognition twice.
  There are three factors to be taken into account when performing binary state recognition.
  They are how to choose $\bm{V}$, how to choose $\bm{Q}$, and how to integrate multiple $\bm{A}$.
  The details of how to choose $\bm{Q}$ are described in \secref{subsec:tips}, and in this section, we describe the integration of $\bm{A}$ using multiple $\bm{V}$ and $\bm{Q}$ results.

  First, if we perform inference using the current state $\bm{V}_{t}$ alone, the result of $\bm{A}$ will fluctuate.
  The variation of $\bm{V}$ is mainly due to two factors: the direction of time $t$, i.e. the viewing direction, and the noise factor, which can differ slightly even for the same viewing direction.
  State recognition for robots is generally performed in a stationary state.
  Of course, it is possible to perform while changing the viewing direction with time $t$, but it is often inconvenient to use such a method because of the motion involved.
  Therefore, in this study, we perform state recognition with some robustness by inferring multiple $\bm{A}$ using multiple $\bm{V}$ with noise added to the current $\bm{V}_{t}$, without changing the viewing direction.

  The state recognition is performed by synthesizing multiple $\bm{A}$ inferred by multiple $\bm{Q}$ (as described in \secref{subsec:tips}) and multiple $\bm{V}$ with noise.
  As we discuss in \secref{subsec:tips}, we restrict questions to ``Is -?'' or ``Does -?'' that will return a binary $\bm{A}$.
  On the other hand, binary questions do not always return Yes or No answers.
  In many cases, the question ``Is the door open?'' can be answered with ``The door is open'' or some other nonsense.
  Therefore, in this study, we ignore the invalid answers other than ``Yes'' or ``No''.
  We calculate the probability of Yes among the samples that return Yes or No, and perform state recognition in a very simple manner, i.e. Yes if the probability exceeds 0.5, and No if otherwise.
}%
{%
  ロボットの動作分岐記述は, 状態出力が"Cat"や"A man with glasses is sitting on a chair"では状態遷移を記述しにくい.
  そこで本研究では, ロボットにおける状態認識という部分に焦点を当て, YesまたはNoによる2値の状態認識のみを扱う.
  状態遷移先が3つある場合は, 2値の状態認識を2回行えば良い.
  ここで, 2値状態認識を行うにあたり, 考慮すべき事象が3つある.
  それは, $\bm{V}$の選び方, $\bm{Q}$の選び方, 複数の$\bm{A}$の統合の仕方である.
  細かい$\bm{Q}$の選び方は\secref{subsec:tips}で述べ, 本節では複数の$\bm{V}$と$\bm{Q}$の結果を用いた$\bm{A}$の統合について述べる.

  まず, 現在の状態$\bm{V}_{t}$であるが, それ単体で推論を行ってしまうと, $\bm{A}$の結果が変動して答えがチラついてしまう.
  $\bm{V}$の変化は, 主に時間$t$の方向, つまり見る方向の違いと, 同じ見え方でも微妙に異なるノイズ要素の2つの要因がある.
  ロボットの動作に向けた状態認識では, 基本的に静止した状態で認識を行うことが多い.
  もちろん時間$t$に伴いロボットの見る方向を変化させながら推論することも可能であるが, 動作が絡むため使いづらいものとなってしまうことが多い.
  そこで本研究では, 姿勢等は変えず, 現在の$\bm{V}_{t}$に対してノイズを加えた複数の$\bm{V}$を用いて複数の$\bm{A}$を推論することで, ある程度ロバストな状態認識を行う.

  \secref{subsec:tips}で述べる複数の$\bm{Q}$と, 前述の複数の$\bm{V}$の組み合わせの推論により得られた$\bm{A}$を総合し, 状態認識を行う.
  \secref{subsec:tips}で述べるが, 本節では``Is -?''や``Do -?''のように, 2値の$\bm{A}$を返す質問のみに限定している.
  一方で, 2値で質問したからと言って, 必ずYesまたはNoで答えが返ってくるわけではない.
  ``Is the door open?''に対して, ``The door is open''という答えや, トンチンカンな答えが返ってくる場合も少なくない.
  そこで本研究では, YesまたはNo以外の答え(Invalid)は無視する.
  YesまたはNoを返すサンプルの中でのYesの確率を計算し, これが0.5を超えればYes, 超えなければNoという非常にシンプルな形で状態認識を行う.
}%

\subsection{Tips for Binary State Recognition Using VQA} \label{subsec:tips}
\switchlanguage%
{%
  In this section, we discuss the important factors in choosing $\bm{Q}$ obtained from the preliminary experiments, and dividing them into the categories of article, state expression, form, and wording.

  \subsubsection{Article}
  The performance of state recognition is affected by changes in the articles ``a'', ``the'', ``this'' and ``that'' added to objects and environments.
  In this study, we perform \secref{subsec:state-recognition} by creating multiple $\bm{Q}$ with different articles, inferring them as a batch, and obtaining multiple $\bm{A}$.

  \subsubsection{State Expression}
  We consider antonyms of adjectives and verbs that express the state.
  For example, ``open'' and ``clean'' are used as adjectives, but $\bm{A}$ may be ambiguous only with each of these adjectives.
  The same applies to verbs such as ``is running'' and ``have''.
  Therefore, we obtain the result more robustly by adding $\bm{Q}$ with the opposite adjective or verb, i.e., ``closed'', ``messy'', ``is not running'', etc., and collecting multiple $\bm{A}$.

  \subsubsection{Form}
  There are various possible forms of $\bm{Q}$.
  Since this study is only concerned with binary state recognition, it is sufficient to obtain a Yes or No answer, so the question is generally ``Is -?'' or ``Are -?''.
  On the other hand, a question ``Does this image look like -?'' is also possible.
  Preliminary experiments have shown that this ``Does -?'' form produces few invalid answers and a high probability of getting a Yes or No response.
  However, the accuracy of recognition decreases (the result is described in \secref{sec:experiment}).
  In addition, the state has the information of ``What'', ``How'', and ``Where''.
  Therefore, it is possible to ask not only binary questions but also questions using ``What'', ``How'', and ``Where'', prepare a set of answers to these questions, and then match them to confirm Yes or No.
  However, this method is not used in this study because it is difficult to prepare answers and to take into account slightly different wording of $\bm{A}$.

  \subsubsection{Wording}
  $\bm{Q}$ should not include words, states, etc. that are not included in the large-scale dataset, and words that the network can recognize should be selected.
  For example, specific brands of drinks or words that describe uncommon situations should not be used.
  Also, there is an important technique here: ask the Image Captioning (IC) question ``What does the image describe?'', and use the obtained words in the $\bm{Q}$.
  For example, a transparent door is generally referred to ``transparent door'' or ``glass door'', but if the word obtained from the IC question is ``window'', $\bm{A}$ often becomes more accurate when using that word.
  It is possible to get better recognition performance by understanding the nature of the network from IC questions.
}%
{%
  本節では予備実験から得られた$\bm{Q}$を選ぶ際に重要な事項を, 冠詞, 状態表現, 形態, 言葉遣いに分けて述べる.

  \subsubsection{冠詞}
  物体や環境に対して, ``a''や``the'', ``this''や``that''の冠詞をつけるが, これらの変化によって状態認識の性能は様々に変化する.
  本研究では, この冠詞を変化させた$\bm{Q}$を複数作成し, バッチとして推論, 複数の$\bm{A}$を得ることで, \secref{subsec:state-recognition}を実行する.

  \subsubsection{状態表現}
  $\bm{Q}$に用いる状態を表現する形容詞や動詞の反対語について考える.
  例えば, 形容詞で言うと``open''や``clean''等があるが, これらだけでは$\bm{A}$が曖昧な場合がある.
  また, 動詞で言うと``is running''や``have''等も同様である.
  そのため, その形容詞や動詞を反対語, つまり``closed''や``messy'', ``is not running''等に変更した$\bm{Q}$も追加し, 複数の$\bm{A}$を得ることで, よりロバストに結果を得ることができる.

  \subsubsection{形態}
  $\bm{Q}$の質問の仕方には様々な形態が考えられる.
  本研究は2値の状態認識のみを行うため, YesまたはNoの答えを得ることができれば良いので, 基本的に質問は``Is -?''や``Are -?''となる.
  一方で, ``Does this image look like -?''等の質問方法も考えられる.
  予備実験から, この質問方法は無効な答えが少なく, 高い確率でYesまたはNoによる返答が得られることがわかっている.
  ただし, ``Is -?''よりも文章が長い分, 認識精度が落ちてしまうという問題もあり, その結果は\secref{sec:experiment}で述べる.
  また, 画像と言語の対応を考えたときに, 状態を示すのは, 何が(``What''), どのように(``How''), どこに(``Where'')あるかという情報である.
  そのため, 2値の質問だけでなく, ``What''や``How'', ``Where''を用いた質問をして, これに対して回答の集合を用意しておき, マッチングしてYesまたはNoを確認するという手段も考えられる.
  しかし, この方法は回答の用意や微妙に異なる単語に対するマッチングが難しいため, 本研究では用いていない.

  \subsubsection{言葉遣い}
  最後に, $\bm{Q}$には大規模データセットに含まれない単語や状態等は入れるべきではなく, ネットワークが認識可能な言葉選びをするべきである.
  例えば, 具体的な飲み物の銘柄や, 一般的ではない状況を現した言葉は用いるべきではない.
  また, ここには重要なテクニックが存在する.
  これは, ``What does the image describe?''というImage Captioningの質問をし, その際に得られた単語を用いるという方法である.
  例えば透明なドアは一般的にtransparent doorやglass doorとして質問するが, その際に得られた単語がwindowである場合, そちらを使った方が精度が高い場合が多い.
  ネットワークの性質を他の質問から理解することでより良い認識性能を引き出すことが可能である.
}%

\section{Experiments} \label{sec:experiment}
\switchlanguage%
{%
  In this study, we mainly present various examples of state recognition that appear in the motion of mobile robots, and discuss their characteristics and performance.
  Many of these state recognitions can be achieved with nearly 100\% accuracy by using point clouds, image processing, supervised learning with image annotations, and conditions of judgements based on individual sensors.
  On the other hand, our method is superior in that it can recognize qualitative values that are difficult to be recognized by the existing state recognition methods, and above all, it can intuitively describe the conditions of judgement by using verbal commands, eliminating the need for individual training and programming.
  Specifically, the robot judges the opening or closing of basic doors, the opening or closing of refrigerator and microwave doors, the opening or closing of transparent doors, the turning on or off of displays, the presence or absence of water, and the cleanliness of the kitchen.
  We also show the results of experiments in which the robot recognizes the opening/closing of a refrigerator and a microwave door at the same time, and in which the robot recognizes the opening/closing of a door and the on/off of a display at the same time.

  In all experiments, the question will be in two forms: ``Does this image look like -?'' (\textbf{Does}) and ``Is -?'' (\textbf{Is}), with the option of four articles \{a, the, this, that\}, and two state expressions that are antonyms, such as \{open, closed\}.
  For $\bm{V}$, five images are generated by RGBShift, which adds randomly selected values from a uniform distribution within the range of [-0.1, 0.1] to each RGB value as noise.
  The results of a total of 40 questions ($4\times2\times5$ excluding the different forms of $\bm{Q}$) are integrated, and the answers are graphed as Correct, Wrong, and Invalid for \textbf{Does} and \textbf{Is}, respectively.
  The percentage of Correct in the sample excluding Invalid is shown, and state recognition is possible when the percentage exceeds 50\%.
  In some experiments, results with multiple wordings are shown.
  The table also shows the rate of the correct response for two kinds of $\bm{V}$ (Img) states (e.g. Img-Open, Img-Closed) when two kinds of $\bm{Q}$ (Ques) state expressions (e.g. Ques-Open, Ques-Closed) are used.
  This allows us to identify which state is difficult to recognize and which state expression in the question is effective for each state.
}%
{%
  本研究では主に移動ロボットの動作において現れる多様な状態認識の例を紹介し, それぞれの特性や性能について議論する.
  これらの状態認識は, ポイントクラウドや画像処理, 画像のアノテーションによる教師あり学習, 個別のセンサを用いた判定により100\%近い正確さで可能な状態認識が多い.
  一方で, これら既存の状態認識では難しい感覚的な, 質的な値の認識も可能である点, 何よりも言語で指令するため直感的に判定条件を記述でき, 個別の学習やプログラムが必要ない点で優れている.
  具体的には, 基本的なドア開閉, 冷蔵庫と電子レンジのドア開閉, 透明なドアの開閉, ディスプレイの電源オンオフ, 水の有無, キッチンの綺麗さ, ロボットの物体把持を判定する.
  また, 冷蔵庫と電子レンジのドア開閉を同時に認識する場合, ドア開閉とディスプレイのオンオフを同時に認識する場合の実験結果も示している.

  なお, 全ての実験で``Does this image look like -?''による質問(\textbf{Does})と``Is -?''による質問(\textbf{Is})という2つの質問形態, \{a, the, this, that\}の4つの冠詞, \{open, closed\}のような反対語を用いる2つの状態表現を用いた$\bm{Q}$による結果を示している.
  $\bm{V}$については, ノイズとしてRGBそれぞれの値に[-0.1, 0.1]の範囲内の一様分布からランダムに選ばれた値を足し込むRGBShiftにより, 5つの画像を作成し用いる.
  形態を除いた$4\times2\times5$の合計40の質問の結果を統合し, \textbf{Does}と\textbf{Is}についてそれぞれ, 回答をCorrect, Wrong, Invalidでグラフとして示している.
  Invalidを除いたサンプルにおけるCorrectの割合を\%表記しており, これが50\%を超えることで認識が可能となる.
  一部の実験では, 複数のwordingを用いた結果を示す場合がある.
  また, 2種類の$\bm{V}$ (Img)の状態(e.g. Img-Open, Img-Closed)について, 2種類の$\bm{Q}$ (Ques)の状態表現(e.g. Ques-Open, Ques-Closed)を用いた場合の正解率もテーブルとして示している.
  これにより, どのような状態の認識が難しく, それぞれの状態についてどのような質問の状態表現が有効であるのかを確認することができる.
}%

\subsection{Is This Door Open?}
\begin{figure}[htb]
  \centering
  \includegraphics[width=1.0\columnwidth]{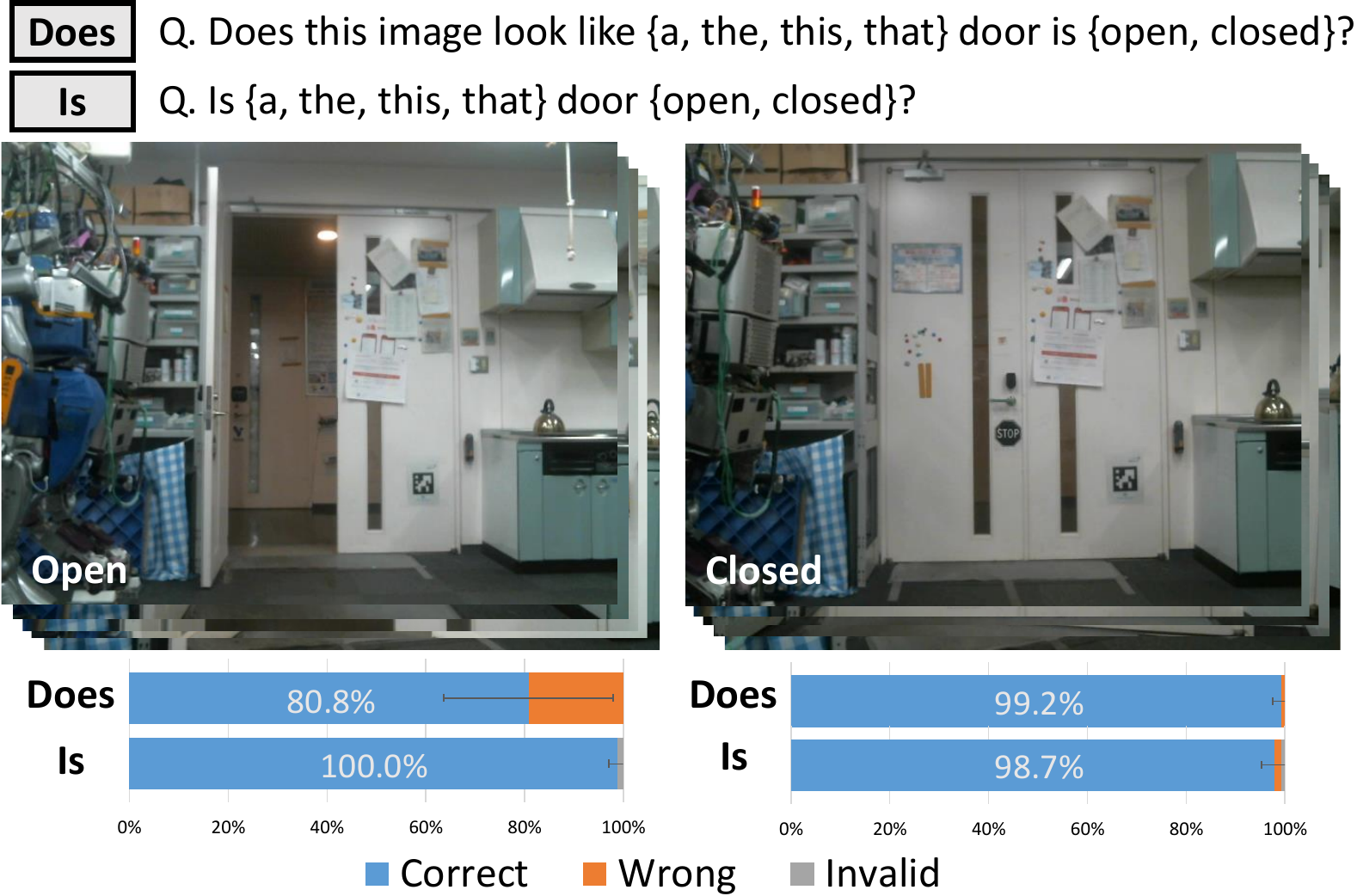}
  \caption{The door state recognition experiment. The figures show the correct, wrong, and invalid rates when changing the article, state expression, and form of question.}
  \label{figure:door}
\end{figure}

\switchlanguage%
{%
  The experimental results on the recognition of the Open/Closed state of a door are shown in \figref{figure:door}.
  Five images for each Open/Closed state are collected from various viewing directions. 
  The mean and variance of the rate of correct responses are shown.
  It can be seen that both forms of questions can recognize the Open/Closed state with high accuracy.
  The recognition accuracy of \textbf{Does} is slightly lower, but that of \textbf{Is} is almost 100\% and the variance is small.
  From \tabref{table:door}, the accuracy is low when the Closed state expression is used in the Open state.
  Therefore, the accuracy is better when only the Open state expression is used.
}%
{%
  \figref{figure:door}にドアの開閉状態(Open or Closed)の認識実験の結果を示す.
  本実験については, 様々な角度から5つずつOpen/Closed状態の画像を取得し, それらの認識率の平均と分散を示している.
  どちらの質問形態でも, 高い精度で開閉状態を認識できていることがわかる.
  \textbf{Does}については多少認識精度が落ちるが, \textbf{Is}についてはほぼ100\%の精度で分散も少ない.
  また, \tabref{table:door}からOpen状態においてClosed状態表現を用いるときの精度のみ多少低いことがわかる.
  逆に, Open状態表現のみを用いればより精度は向上する.
}%

\begin{table}[htb]
  \centering
  \caption{Performance difference regarding image state (Img-Open/Closed) and state expression of the question (Ques-Open/Closed) for door state recognition experiment.}
  \begin{tabular}{l||cc|c}
    &  Ques-Open& Ques-Closed & Total \\ \hline\hline
    Img-Open  & 0.983 & 0.824 & 0.904 \\
    Img-Closed & 0.992 & 0.988 & 0.990 \\ \hline
    Total & 0.987 & 0.906 & \\
  \end{tabular}
  \label{table:door}
\end{table}

%

\subsection{Is  This Refrigerator / Microwave Door Open?}
\begin{figure}[htb]
  \centering
  \includegraphics[width=1.0\columnwidth]{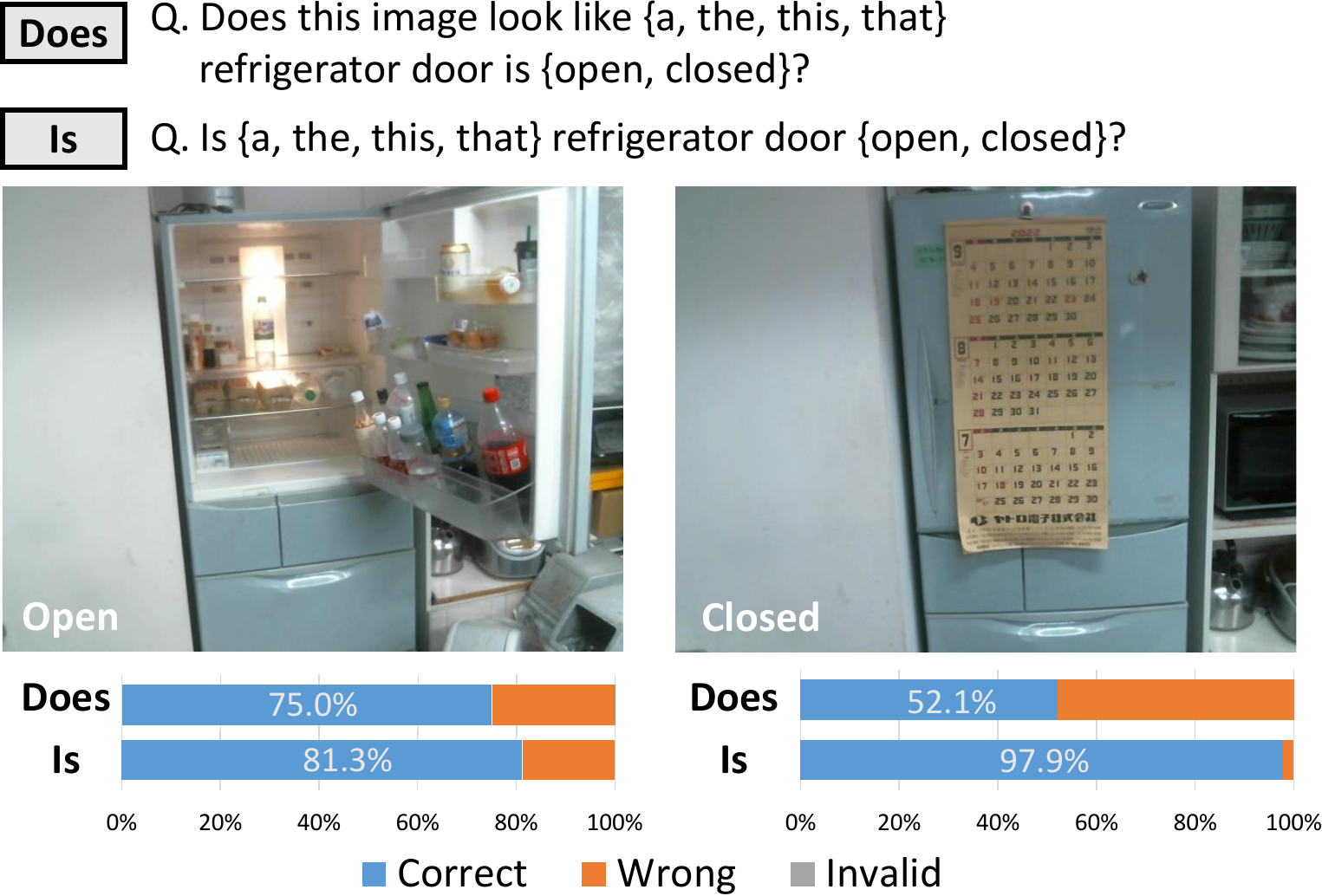}
  \caption{The refrigerator door state recognition experiment.}
  \label{figure:refrigerator}
\end{figure}

\begin{figure}[htb]
  \centering
  \includegraphics[width=1.0\columnwidth]{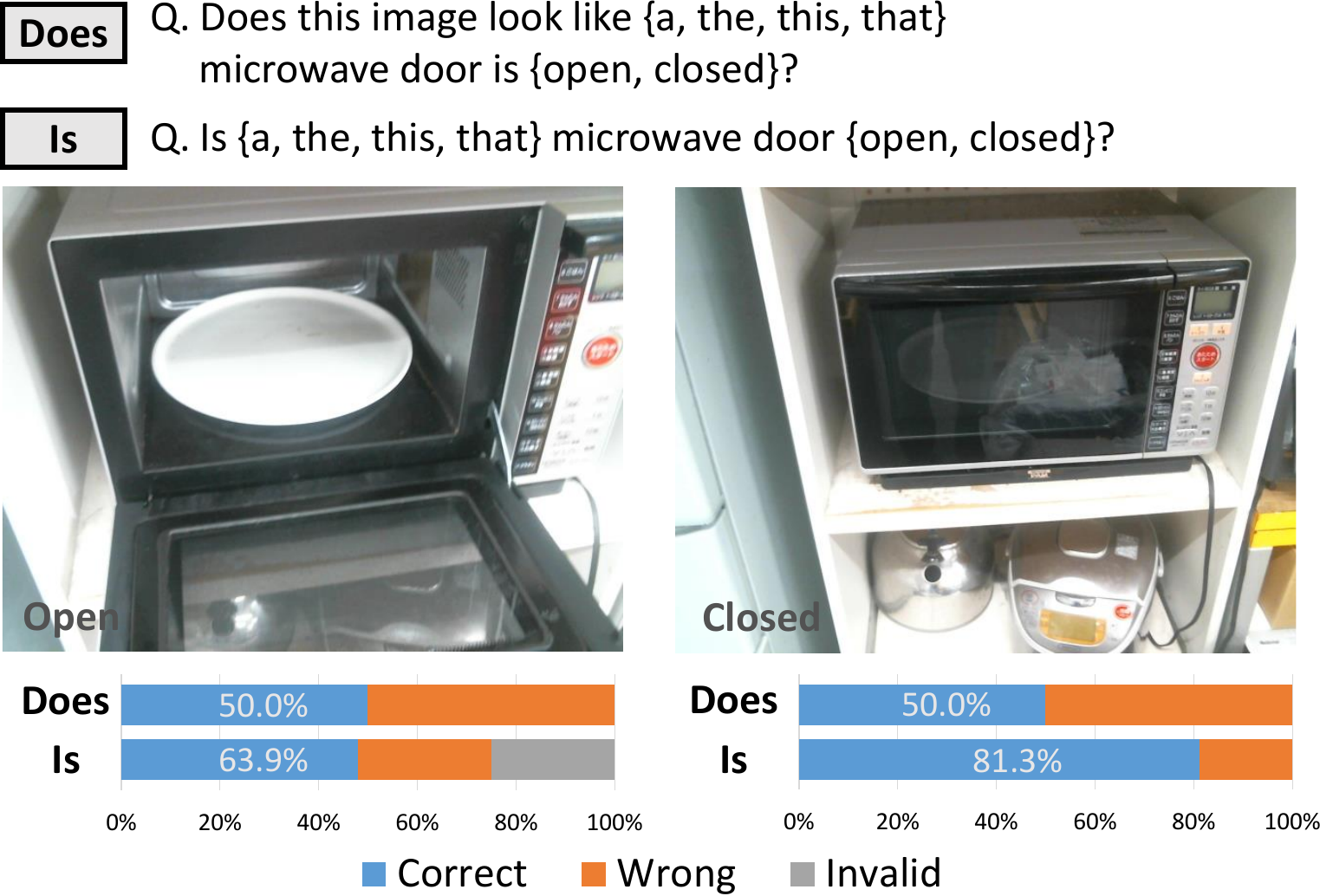}
  \caption{The microwave door state recognition experiment.}
  \label{figure:microwave}
\end{figure}

\switchlanguage%
{%
  The experimental results on the recognition of the Open/Closed state of a refrigerator door are shown in \figref{figure:refrigerator}.
  By using the form of \textbf{Is}, the Open/Closed state of the door is recognized with high accuracy.
  On the other hand, though \textbf{Does} can also recognize the Open/Closed state of the door, its accuracy is low.
  From \tabref{table:refrigerator}, we can see that the accuracy of questions using the correct state expression for the state is good (i.e. using the word ``open'' when the door is actually open), but the accuracy of questions using the opposite state expression is poor.
  Therefore, it is important to integrate the results of both state expressions.

  The experimental results on the recognition of the Open/Closed state of a microwave door are shown in \figref{figure:microwave}.
  By using the form of \textbf{Is}, the Open/Closed state of the door is recognized.
  On the other hand, it is difficult to recognize the Open/Closed state with \textbf{Does}.
  From \tabref{table:microwave}, we can see that the accuracy of questions using the correct state expression for the state is good, but the accuracy of questions using the opposite state expression is poor, as in the refrigerator experiment.
}%
{%

  \figref{figure:refrigerator}に冷蔵庫ドアの開閉状態認識実験の結果を示す.
  \textbf{Is}の形態を利用することで, 高い精度で開閉状態を認識できている.
  一方, \textbf{Does}でも開閉認識認識はできるものの, その精度は低い.
  また, \tabref{table:refrigerator}から, Open/Closed状態に対して, 同様の状態表現を用いた質問の精度は良いが, 異なる状態表現を用いた質問の精度は悪いことがわかる.
  そのため, 2種類の状態表現どちらも大事であり, それらの結果を統合する必要がある.

  \figref{figure:microwave}に電子レンジドアの開閉状態認識実験の結果を示す.
  \textbf{Is}の形態を利用することで, 開閉状態を認識できている.
  一方, \textbf{Does}では開閉認識認識は難しい.
  また, \tabref{table:microwave}から冷蔵庫同様, Open/Closed状態の画像に対して, 同様の状態表現を用いた質問の精度は良いが, 異なる状態表現を用いた質問の精度は悪いことがわかる.
}%

\begin{table}[htb]
  \centering
  \caption{Performance difference regarding image state (Img-Open/Closed) and state expression of the question (Ques-Open/Closed) for the refrigerator door state recognition experiment.}
  \begin{tabular}{l||cc|c}
    &  Ques-Open& Ques-Closed & Total \\ \hline\hline
    Img-Open  & 0.979 & 0.583 & 0.781 \\
    Img-Closed & 0.500 & 1.000 & 0.750 \\ \hline
    Total & 0.740 & 0.792 & \\
  \end{tabular}
  \label{table:refrigerator}
\end{table}

\begin{table}[htb]
  \centering
  \caption{Performance difference regarding image state (Img-Open/Closed) and state expression of the question (Ques-Open/Closed) for the microwave door state recognition experiment.}
  \begin{tabular}{l||cc|c}
    &  Ques-Open& Ques-Closed & Total \\ \hline\hline
    Img-Open  & 0.958 & 0.028 & 0.560 \\
    Img-Closed & 0.313 & 1.000 & 0.656 \\ \hline
    Total & 0.645 & 0.583 & \\
  \end{tabular}
  \label{table:microwave}
\end{table}

%
%

\subsection{Is This Transparent Door Open?}
\begin{figure}[htb]
  \centering
  \includegraphics[width=1.0\columnwidth]{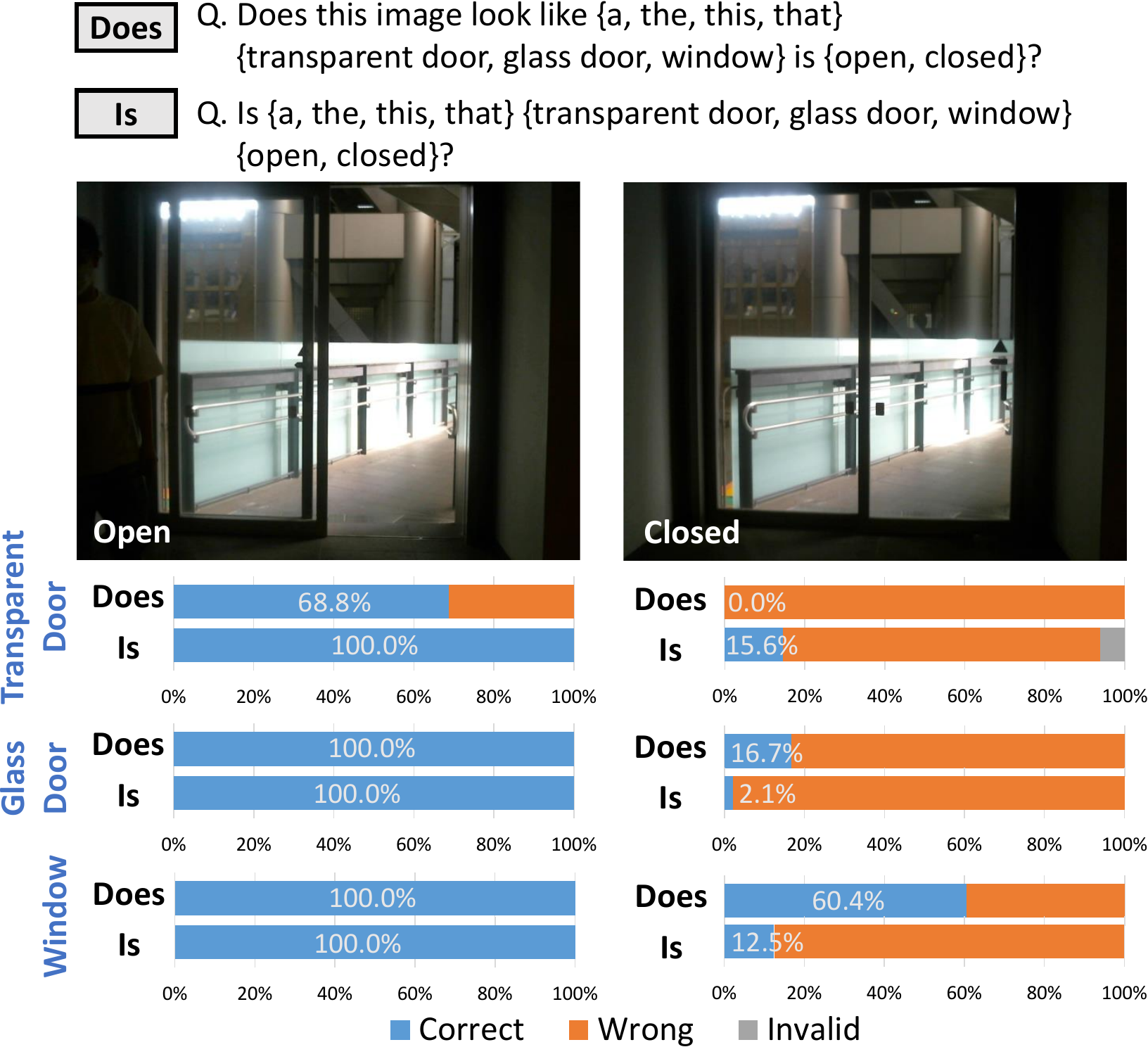}
  \caption{The transparent door state recognition experiment.}
  \label{figure:transparent}
\end{figure}

\switchlanguage%
{%
  The experimental results on the recognition of the Open/Closed state of a transparent door are shown in \figref{figure:transparent}.
  As wording, we use the direct expression ``transparent door'', as well as ``glass door'' and ``window'' obtained from IC.
  Note that the result of IC in the Open state is ``a view of a building through a glass door'', and that of IC in the Closed state is ``a view through a window of a parking garage''.
  Since the door is transparent, the door is always recognized as Open and cannot be recognized as Closed, which results in a significantly worse recognition accuracy for the Closed state.
  This property can also be seen from \tabref{table:transparent}, which shows that the recognition accuracy is high only for the Open state.
  On the other hand, the recognition accuracy for ``glass door'' is somewhat better than that for ``transparent door''.
  Also, using ``window'', which is the result from IC in the Closed state, allows us to recognize the Closed state as Closed, which \textbf{Does} succeeds in.
  When ``wording'' is set to ``window'' and only the Closed state expression is used in \textbf{Does}, the Open/Closed state is recognized with 100\% accuracy.
  It is important to adjust the wording and the state expression according for the best performance.
}%
{%
  \figref{figure:transparent}に透明なドアの開閉状態認識実験の結果を示す.
  ここでは, wordingとして, 純粋な表現である``transparent door'', そして, ICから得られた``glass door''と``window''を用いる.
  なお, Open状態におけるICの結果は``a view of a building through a glass door'', Closed状態におけるICの結果は``a view through a window of a parking garage''である.
  ドアが透明であるがゆえ, Open状態でもClosed状態でも, 常にOpenであると認識し, Closedであることが認識できないため, Closed状態の画像に対する認識精度が著しく悪くなってしまっている.
  一方その中でも, ``glass door''を使った場合は``transparent door''を使った場合よりも多少認識精度が向上する.
  また, Closed状態におけるICの結果である``window''を用いることで, Closed状態をClosedであると認識できるようになり, \textbf{Does}では状態認識に成功している.
  その性質は\tabref{table:transparent}からも見ることができ, Open状態に対する認識精度だけ高いことがわかる.
  一方, \textbf{Does}において, wordingを``window''とし, Closedの状態表現のみを用いることで100\%の精度で開閉状態が認識できている.
  wordingやstate expressionを認識したい対象に合わせて調整することが重要であることがわかる.
}%

\begin{table}[htb]
  \centering
  \caption{Performance difference regarding image state (Img-Open/Closed) and state expression of the question (Ques-Open/Closed) for the transparent door state recognition experiment.}
  \begin{tabular}{l||cc|c}
    &  Ques-Open& Ques-Closed & Total \\ \hline\hline
    Img-Open  & 1.000 & 0.896 & 0.948 \\
    Img-Closed & 0.092 & 0.264 & 0.179 \\ \hline
    Total & 0.551 & 0.580 & \\
  \end{tabular}
  \label{table:transparent}
\end{table}

\subsection{Is This Display On?}
\begin{figure}[htb]
  \centering
  \includegraphics[width=1.0\columnwidth]{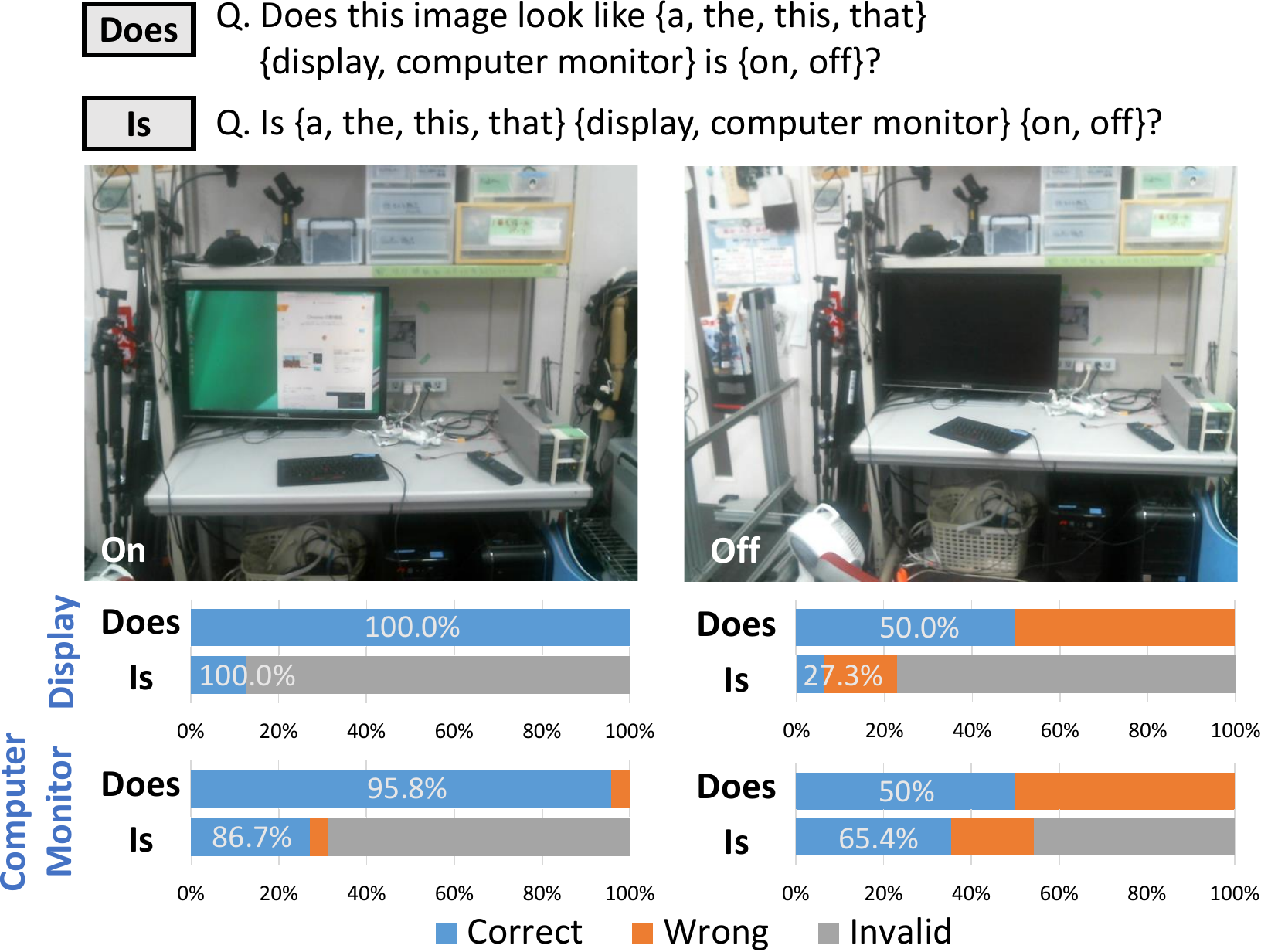}
  \caption{The display state recognition experiment.}
  \label{figure:display}
\end{figure}

\switchlanguage%
{%
  The experimental results on the recognition of the On/Off state of a display are shown in \figref{figure:display}.
  Here, as wording, we use ``display'', as well as ``computer monitor'' obtained from IC.
  Note that the result of IC in the On state is ``a computer monitor sitting on top of a desk'', and that of IC in the Off state is ``a desk with a computer monitor on top of it''.
  In the case of using ``display'', high recognition accuracy is obtained in the On state, but recognition in the Off state proved to be difficult.
  This property can also be seen from \tabref{table:display}, which shows high recognition accuracy for the On state and low recognition accuracy for the Off state.
  On the other hand, using ``computer monitor'', which is the result of IC, succeeds in recognizing both states, especially in the case of \textbf{Is}.
  Note that when only the Off state expression is used, the accuracy is higher for both On and Off states, and it is possible to achieve a high recognition accuracy.
}%
{%
  \figref{figure:display}にディスプレイ状態(On/Off)の認識実験の結果を示す.
  ここでは, wordingとして, ``display'', そして, ICから得られた``computer monitor''を用いる.
  なお, On状態におけるICの結果は``a computer monitor sitting on top of a desk'', Off状態におけるICの結果は``a desk with a computer monitor on top of it''である.
  ``display''を用いた場合は, On状態では高い認識精度が得られるが, Off状態の認識が難しい.
  一方, ICの結果である``computer monitor''を用いることで, 特に\textbf{Is}の場合で状態認識に成功している.
  その性質は\tabref{table:display}からも見ることができ, On状態に対する認識精度が高く, Off状態の認識精度が低い.
  一方, 質問の状態表現をOffとした場合は, 状態がOnでもOffでも精度が高く, Offの状態表現だけを用いれば, \textbf{Does}でも\textbf{Is}でも100\%近い認識精度を達成可能であった.
}%

\begin{table}[htb]
  \centering
  \caption{Performance difference regarding image state (Img-On/Off) and state expression of the question (Ques-On/Off) for the display state recognition experiment.}
  \begin{tabular}{l||cc|c}
    &  Ques-On& Ques-Off & Total \\ \hline\hline
    Img-On  & 1.000 & 0.937 & 0.966 \\
    Img-Off & 0.053 & 0.856 & 0.511 \\ \hline
    Total & 0.514 & 0.892 & \\
  \end{tabular}
  \label{table:display}
\end{table}

\subsection{Is Water Running in This Sink?}
\begin{figure}[htb]
  \centering
  \includegraphics[width=1.0\columnwidth]{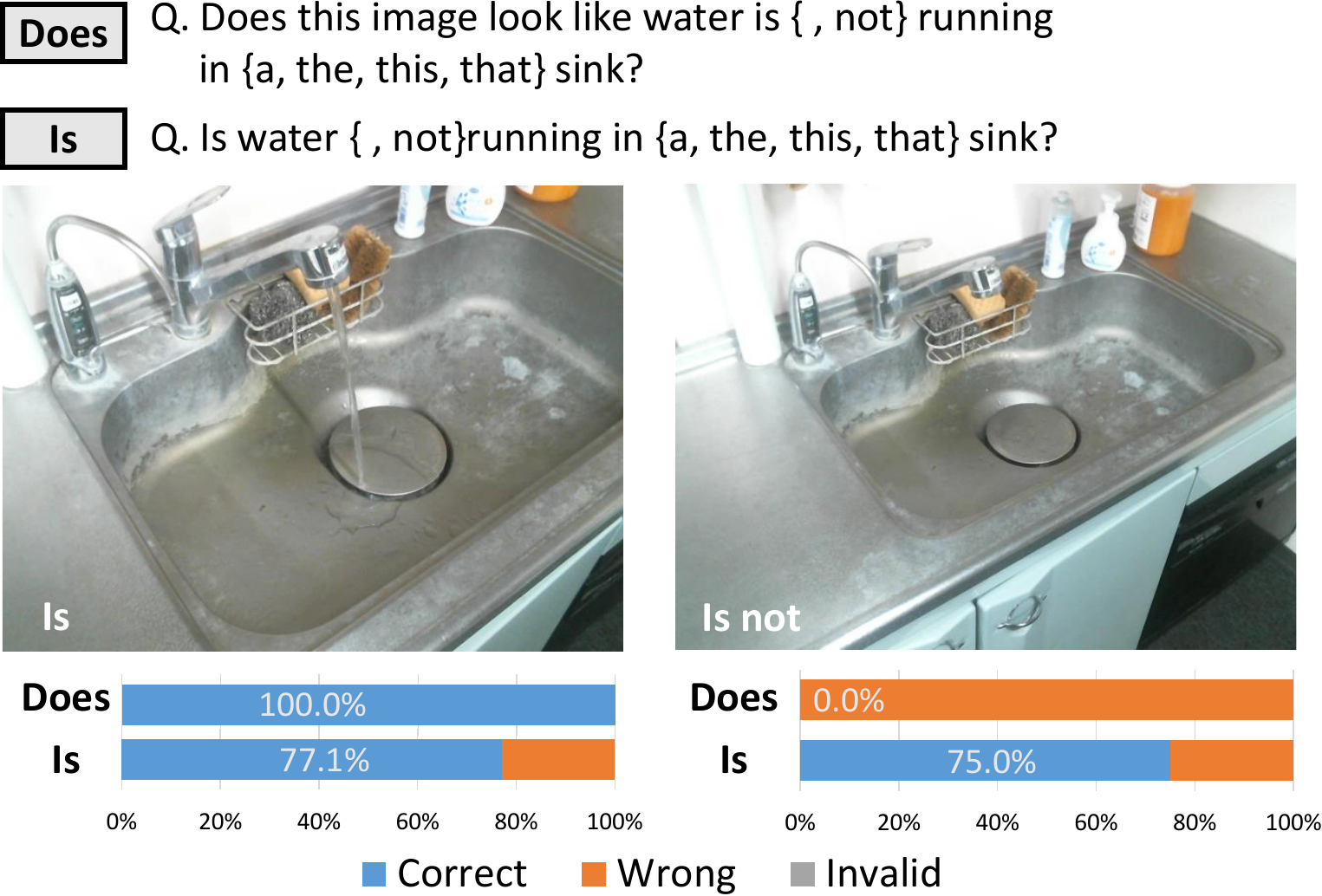}
  \caption{The water state recognition experiment.}
  \label{figure:water}
\end{figure}

\switchlanguage%
{%
  The experimental results on the recognition of the Is/IsNot state (presence or absence) of water running into the sink are shown in \figref{figure:water}.
  By using the form of \textbf{Is}, the Is/IsNot state of water can be recognized.
  From \tabref{table:water}, we can see that the recognition accuracy is high when water is running, but the accuracy is low when water is not running.
  On the other hand, when only the \textbf{Is} state expression is used, the accuracy is high for both Is and IsNot states.
  Also, the state can be recognized with 100\% accuracy by using only the \textbf{Is} form and Is state expression.
}%
{%
  \figref{figure:water}にシンクに流れる水の有無(Is or IsNot)の認識実験の結果を示す.
  \textbf{Is}の形態を利用することで, 水の有無を認識できている.
  また, \tabref{table:water}から, 水が流れているときの認識精度は高いものの, 水が流れていないときも流れていると錯覚し, 精度が低くなっていることがわかる.
  一方, 質問の状態表現をIsとした場合は, 状態がIsでもIsNotでも精度が高く, 形態\textbf{Is}においてのIs状態表現のみを用いることで100\%の精度で水の有無が認識できている.
}%

\begin{table}[htb]
  \centering
  \caption{Performance difference regarding image state (Img-Is/IsNot) and state expression of the question (Ques-Is/IsNot) for the water state recognition experiment.}
  \begin{tabular}{l||cc|c}
    &  Ques-Is& Ques-IsNot & Total \\ \hline\hline
    Img-Is & 1.000 & 0.563 & 0.781 \\
    Img-IsNot & 0.500 & 0.271 & 0.380 \\ \hline
    Total & 0.750 & 0.417 & \\
  \end{tabular}
  \label{table:water}
\end{table}

\subsection{Is This Kitchen Clean?}
\begin{figure}[htb]
  \centering
  \includegraphics[width=1.0\columnwidth]{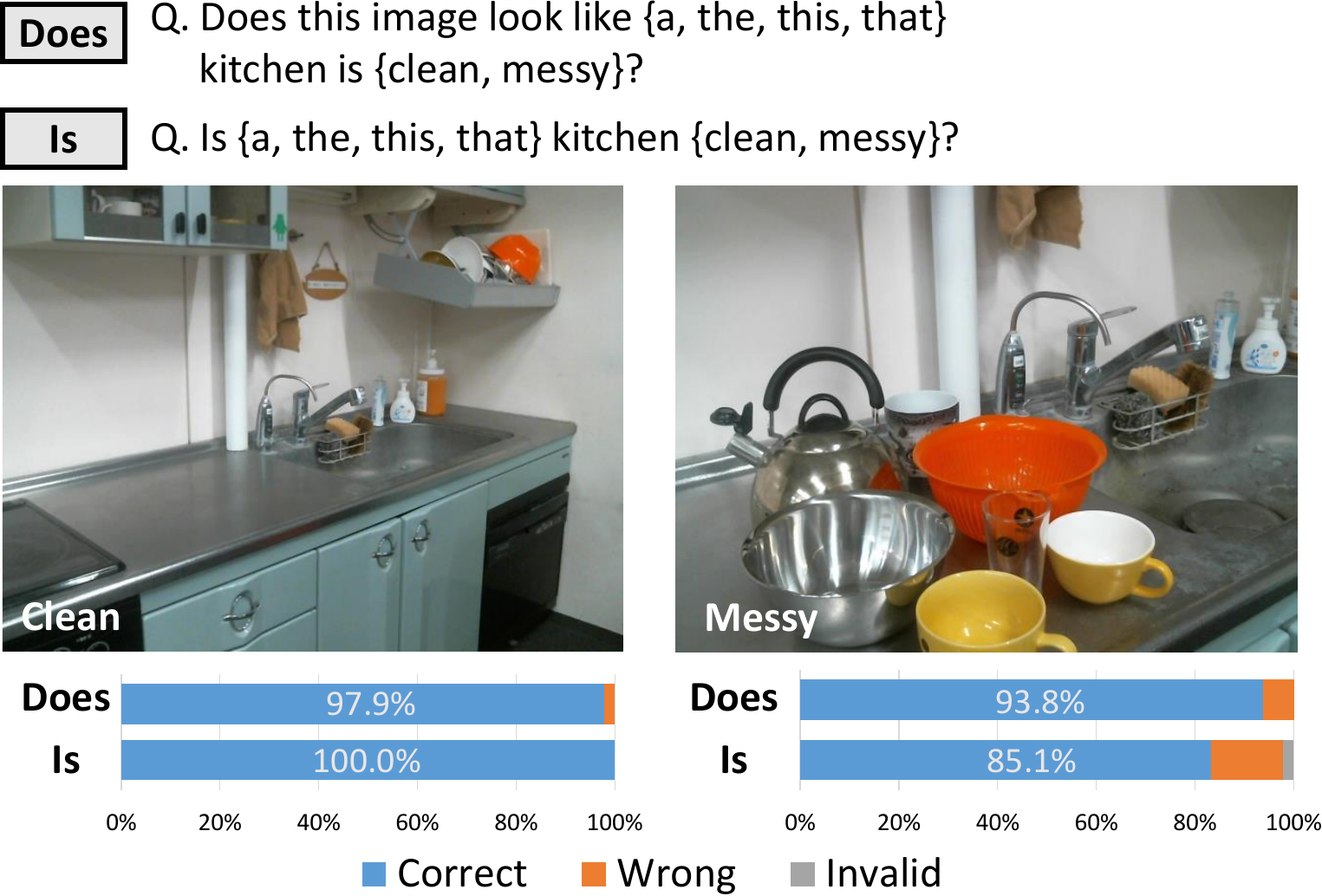}
  \caption{The kitchen state recognition experiment.}
  \label{figure:kitchen}
\end{figure}

\switchlanguage%
{%

  The experimental results on the recognition of the Clean/Messy state of a kitchen are shown in \figref{figure:kitchen}.
  Both forms of the question can correctly recognize the kitchen state.
  From \tabref{table:kitchen}, we can see the high recognition accuracy, but it can also be seen that in the Messy state, the accuracy is somewhat lower when the Clean state expression is used.
  Qualitative state recognition is possible, and general human notion obtained from a large-scale dataset can be used for state recognition.
}%
{%
  \figref{figure:kitchen}にキッチン状態(Clean or Messy)の認識実験の結果を示す.
  どちらの質問形態でも, 正しく状態を認識できている.
  また, \tabref{table:kitchen}からその高い精度がわかるが, Messy状態において, Clean状態表現を用いた場合は多少精度が低いこともわかる.
  このように, 質的な状態認識も可能であり, 大規模データセットから得られた一般的な人間の感覚を状態認識に用いることができる.
}%

\begin{table}[htb]
  \centering
  \caption{Performance difference regarding image state (Img-Clean/Messy) and state expression of the question (Ques-Clean/Messy) for the kitchen state recognition experiment.}
  \begin{tabular}{l||cc|c}
    &  Ques-Clean& Ques-Messy & Total \\ \hline\hline
    Img-Clean  & 1.000 & 0.979 & 0.990 \\
    Img-Messy & 0.787 & 1.000 & 0.895 \\ \hline
    Total & 0.895 & 0.990 & \\
  \end{tabular}
  \label{table:kitchen}
\end{table}

\subsection{Is This Refrigerator Door Open? \& Is This Microwave Door Open?} \label{subsec:combination}
\begin{figure}[htb]
  \centering
  \includegraphics[width=1.0\columnwidth]{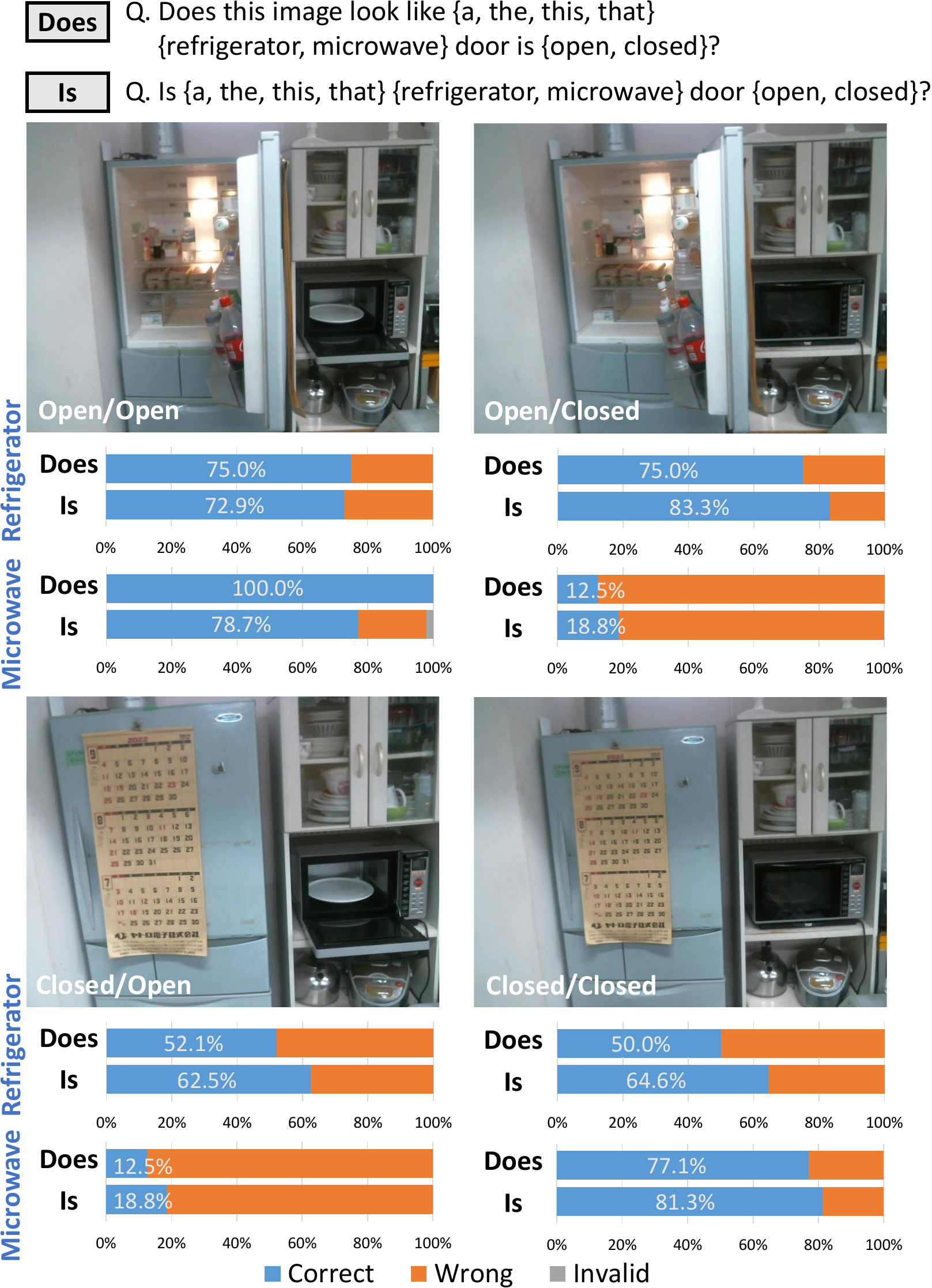}
  \caption{The simultaneous recognition experiment of the refrigerator and microwave door states.}
  \label{figure:combination}
\end{figure}

\switchlanguage%
{%
  The experimental results on the simultaneous recognition of the Open/Closed states of a refrigerator and microwave door are shown in \figref{figure:combination}.
  In both forms, the recognition accuracy is high when the two doors are in the same Open or Closed states, but low when the two doors are in different states.
  For \textbf{Does} in particular, the recognition of the microwave door state fails when the doors are in different Open/Closed states, while the recognition of the refrigerator door state succeeds in all cases.
  This is because the answers are strongly affected by the state of the refrigerator door regardless of the state of the microwave door.
  Although this is a case where the area of the refrigerator is larger than that of the microwave in the image, experiments have shown that the larger area of the microwave in the image strongly affects the recognition of the refrigerator door state to become the microwave door state.
  Thus, it is found that the recognition of two semantically similar states is prone to interference.
}%
{%
  \figref{figure:combination}に, 冷蔵庫と電子レンジがそれぞれ存在する状態において, それぞれのドアの開閉状態認識実験の結果を示す.
  どちらの形態でも, 2つのドアがOpen状態な場合, Closed状態な場合の認識精度は高いが, 2つの状態が異なる場合の認識精度が低いことがわかる.
  特に, \textbf{Does}について, 冷蔵庫の状態認識は全て成功しているのに対して, 両者の開閉状態が異なる場合に, 電子レンジの状態認識が失敗している.
  これは, 電子レンジの開閉状態に関わらず, 回答が冷蔵庫の開閉状態に引っ張られてしまっていることが理由である.
  なお, 電子レンジに比べ冷蔵庫が閉める画像領域が大きい場合はこのような結果となるが, 実験から, 電子レンジを大きく映すことで, 電子レンジの開閉状態に引っ張られるようになることがわかっている.
  このように, 質的に似た2つの状態認識は干渉を起こしやすいことがわかった.
}%

\subsection{Is This Door Open? \& Is This Display On?}
\begin{figure}[htb]
  \centering
  \includegraphics[width=1.0\columnwidth]{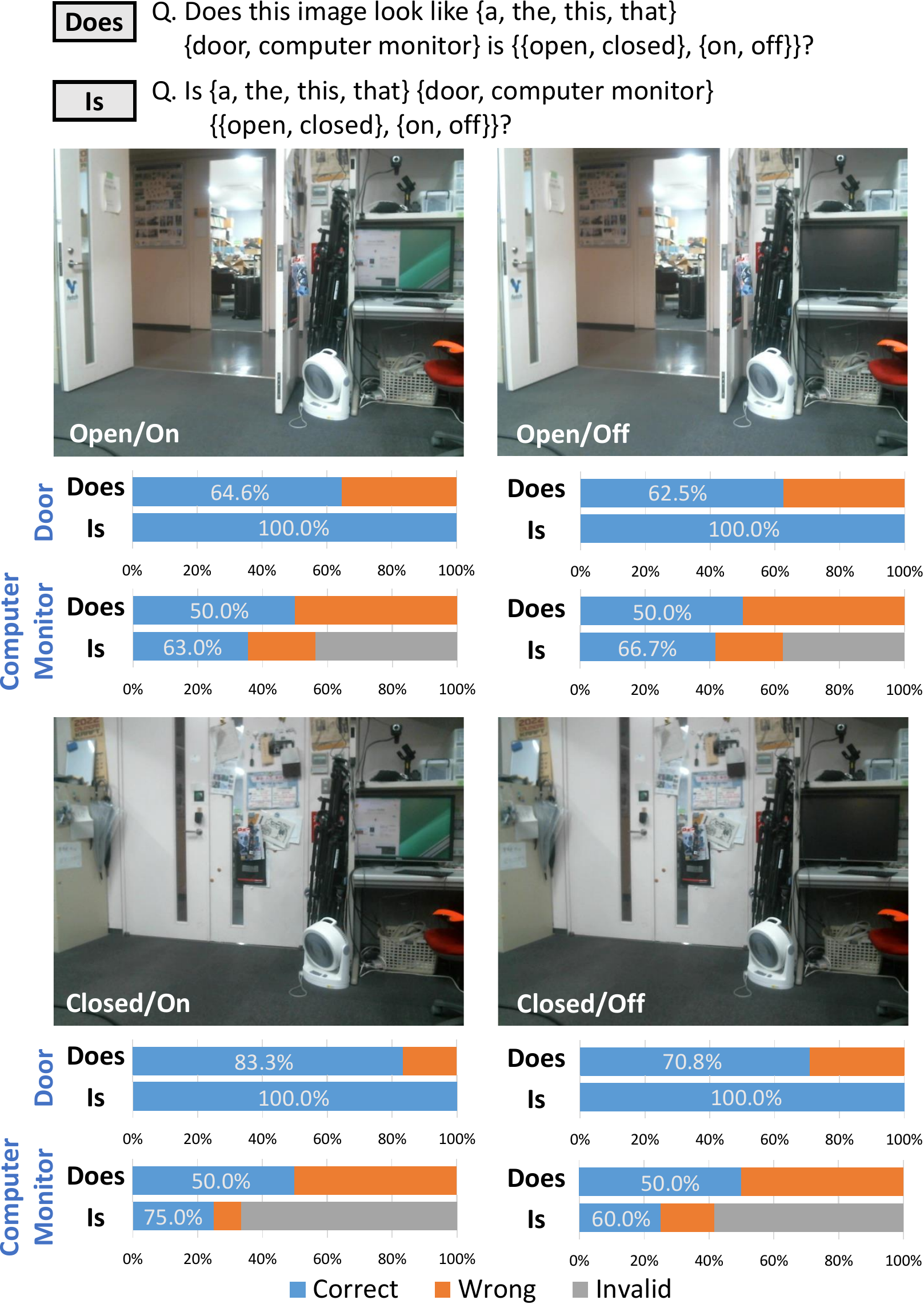}
  \caption{The simultaneous recognition experiment of the door and display states.}
  \label{figure:combination3}
\end{figure}

\switchlanguage%
{%
  The experimental results on the simultaneous recognition of the Open/Closed states of a door and the On/Off states of a display are shown in \figref{figure:combination3}.
  It can be seen that \textbf{Is} correctly recognizes the states for all four cases.
  This is different from the result of \secref{subsec:combination}, and the recognition of two semantically different states is less likely to cause interference.
}%
{%
  \figref{figure:combination3}に, doorとcomputer monitorがそれぞれ存在する状態において, それぞれのドア開閉状態とディスプレイのオンオフ状態の認識実験の結果を示す.
  \textbf{Is}では, 4つ全ての状態について正しく状態認識ができていることがわかる.
  これは\secref{subsec:combination}の結果とは異なる.
  つまり, 質的に異なる2つの状態認識は干渉を起こしにくいことがわかった.
}%

\begin{table}[htb]
  \centering
  \caption{The Correct and Invalid rates of all collected answers according to the form and article of the question.}
  \begin{tabular}{l|cccccc}
    & \textbf{Does} & \textbf{Is} & a & the & this & that\\ \hline
    Correct Rate & 0.698 & 0.804 & 0.714 & 0.738 & 0.756 & 0.707 \\
    Invalid Rate & 0.000 & 0.124 & 0.047 & 0.079 & 0.037 & 0.084 \\
  \end{tabular}
  \label{table:article}
\end{table}

\section{Discussion} \label{sec:discussion}
\switchlanguage%
{%
  Summarizing the results obtained from the experiments, we found the following.
  \begin{itemize}
    \item Recognition performance varies greatly depending on the form, the article, and the state expression of the question, but various states such as Open/Closed, On/Off, Is/IsNot, etc. can be recognized depending on the combination.
    \item Recognition accuracy can be improved by using wording that is recognized by Image Captioning.
    \item It is possible to judge qualitative states such as "Clean" or "Messy".
    \item Two semantically similar states are likely to cause interference, while semantically different states are unlikely to cause interference.
  \end{itemize}
  By using these findings, we can efficiently create various state recognizers.

  In addition, we show the change in recognition performance from the change in the form of the question and the article in \tabref{table:article}, for all experiments we have conducted so far.
  Here, the Correct Rate is the rate of Correct among the Correct and Wrong samples, and the Invalid Rate is the rate of Invalid in all samples.
  As a whole, it can be seen that the Correct Rate of \textbf{Is} is higher than that of \textbf{Does}, and at the same time, the number of Invalid responses of \textbf{Is} is also higher.
  Regarding the articles, the performance of ``this'' is found to be the best, since the order of correct responses is this$>$the$>$a$>$that, and the order of Invalid responses is this$<$a$<$the$<$that.
  The performance varies to some extent depending on the model, but what is particularly important is that the performance can be adjusted depending on the form of the question and the article that is used.

  Finally, we show an automatic patrol experiment in a supplementary video.
  By using the proposed state recognition on the mobile robot Fetch, the robot can close the refrigerator door when it is open, verify that the cabinet door is closed, and exit the room when the door is open.
}%
{%
  実験から得られた結果をまとめると, 以下のことがわかった.
  \begin{itemize}
    \item 質問の形態や冠詞, 状態表現によって認識性能が大きく変化するが, その組み合わせ次第ではOpen/ClosedやOn/Off, Is/IsNotなど様々な状態認識が可能
    \item ICにより認識された単語を用いることで認識精度が向上
    \item 綺麗・汚いのような質的な2値状態判断も可能.
    \item 意味的に似た2つの状態認識は干渉を起こしやすく, 意味的に異なる状態認識は干渉を起こしにくい.
  \end{itemize}
  これらの知見を利用することで, 様々な状態認識を効率的に作成することができる.

  また, これまで行った全ての実験における質問の形態と冠詞変化による性能変化を\tabref{table:article}に示す.
  ここで, Correct Rateとは, CorrectまたはWrongのサンプル中のCorrectの割合, Invalid Rateは全サンプル中のInvalidの割合である.
  全体として, 例外は存在はするものの, \textbf{Is}の方が\textbf{Does}に比べて正答率が高いと同時に, Invalidな回答も多いことがわかる.
  冠詞はthis$>$the$>$a$>$thatの順で正答率が高く, this$<$a$<$the$<$thatの順でInvalidな回答が少ないため, ``this''の性能が最も良いことがわかった.
  これらの性能はモデルによってもある程度変化するが, 特に重要なことは, 質問の形態や冠詞によってその性能を調整できるという点である.
}%

\section{CONCLUSION} \label{sec:conclusion}
\switchlanguage%
{%
  In this study, we performed binary state recognition by a robot using Visual Question Answering in a Pre-Trained Vision-Language Model.
  The robot can probabilistically recognize the states from answers obtained from multiple images and questions.
  The accuracy can be improved by changing the articles, question forms, and state expressions, and by utilizing the wording obtained from Image Captioning.
  No re-training is required, the conditional branching can be described intuitively in the spoken language, and the recognition performance can be easily adjusted.
  The use of appropriate questions enables the robot to recognize not only the opening and closing of microwave and refrigerator doors, but also the opening and closing of transparent glass doors, whether the kitchen is clean or messy, and whether water is running or not in the sink, which have been challenging so far.
  We hope that this idea will revolutionize the recognition behavior of various robots by making it more intelligent, easy, and adjustable.
}%
{%
  本研究では, 事前訓練済みのVision-Language ModelにおけるVisual Question Answeringを用いたロボットによる状態認識を行った.
  複数の画像と質問から得られた答えから確率的に2値状態認識が可能である.
  この際, 冠詞の変化や質問形態の変化, 反対語の利用による状態表現やキャプショニングによって得られた単語の利用などにより, その精度を上げることができる.
  一切の再学習等は必要なく, 言語で直感的に条件分岐を記述可能かつ, その認識性能も調整しやすい.
  適切な質問の利用は, 電子レンジや冷蔵庫のドアの開閉だけでなく, 透明なガラスドアの開閉認識やキッチンの綺麗/汚い, 水が出ているかの認識まで可能とする.
  本研究が様々なロボットの認識行動をより知的かつ容易にし, 革新することを期待する.
}%

{
  \bibliographystyle{IEEEtran}
  \bibliography{main}
}

\end{document}